\documentclass[conference]{IEEEtran}
\IEEEoverridecommandlockouts
\usepackage{cite}
\usepackage{amsmath}
\usepackage{graphicx}
\usepackage{textcomp}
\usepackage{xcolor}
\usepackage{algpseudocode}
\usepackage{booktabs, tabularx}
\usepackage{multirow}
\usepackage{url}
\usepackage{balance}
\usepackage{hyperref}

\def\BibTeX{{\rm B\kern-.05em{\sc i\kern-.025em b}\kern-.08em
    T\kern-.1667em\lower.7ex\hbox{E}\kern-.125emX}}
\begin{document}

\title{Edge AI in Highly Volatile Environments: Is Fairness Worth the Accuracy Trade-off?}

\author{
\IEEEauthorblockN{ Obaidullah Zaland \IEEEauthorrefmark{1}, Feras M. Awaysheh \IEEEauthorrefmark{1}, Sawsan Al Zubi \IEEEauthorrefmark{3}, Abdul Rahman Safi \IEEEauthorrefmark{4}, and Monowar Bhuyan \IEEEauthorrefmark{1} }
\IEEEauthorblockA{\IEEEauthorrefmark{1}\textit{Department of Computing Science},
\textit{Umeå University},
Umeå, SE-90187, Sweden,\\
 \{obaidullah.zaland, feras.awaysheh, monowar.bhuyan\}@umu.se}

\IEEEauthorblockA{\IEEEauthorrefmark{3}\textit{University of Santiago de Compostela} \\
Santiago de Compostela, Spain,
sawsan.alzubi@usc.es}

\IEEEauthorblockA{\IEEEauthorrefmark{4}\textit{Kabul University},
Kabul, Afghanistan\\
abrahman.safi@gmail.com}
\thanks{This work was partially supported by the Wallenberg AI, Autonomous Systems and Software Program (WASP) funded by the Knut and Alice Wallenberg Foundation via the WASP NEST project “Intelligent Cloud Robotics for Real-Time Manipulation at Scale.”}
}

\maketitle

\begin{abstract}

Federated learning (FL) has emerged as a transformative paradigm for edge intelligence, enabling collaborative model training while preserving data privacy across distributed personal devices. However, the inherent volatility of edge environments, characterized by dynamic resource availability and heterogeneous client capabilities, poses significant challenges for achieving high accuracy and fairness in client participation. This paper investigates the fundamental trade-off between model accuracy and fairness in highly volatile edge environments. This paper provides an extensive empirical evaluation of fairness-based client selection algorithms such as RBFF and RBCSF against random and greedy client selection regarding fairness, model performance, and time, in three benchmarking datasets (CIFAR10, FashionMNIST, and EMNIST). This work aims to shed light on the fairness-performance and fairness-speed trade-offs in a volatile edge environment and explore potential future research opportunities to address existing pitfalls in \textit{fair client selection} strategies in FL. Our results indicate that more equitable client selection algorithms, while providing a marginally better opportunity among clients, can result in slower global training in volatile environments\footnote{The code for our experiments can be found at \url{https://github.com/obaidullahzaland/FairFL_FLTA}}. 
\end{abstract}

\begin{IEEEkeywords}
Federated learning, Edge Intelligence, Fairness, Client Selection, Responsible FL
\end{IEEEkeywords}

\section{Introduction}

The proliferation of Internet of Things (IoT) and edge computing infrastructure \cite{awaysheh2021security} has fundamentally transformed how we approach machine learning in distributed environments. Edge intelligence, which brings computational capabilities closer to data sources, has emerged as a critical paradigm for enabling real-time decision-making while addressing privacy concerns and bandwidth limitations \cite{shi2016edge}. Within this context, federated learning (FL) \cite{zaland2025one,mcmahan2017communication} has gained significant traction as a privacy-preserving approach that enables multiple edge devices to collaboratively train machine learning (ML) models without sharing raw data \cite{zalandheterogeneity}.

However, the deployment of FL in edge environments presents unique challenges that traditional centralized learning approaches do not encounter. Edge devices are inherently heterogeneous, exhibiting significant variations in computational capabilities, communication bandwidth, energy constraints, and data quality \cite{li2020federated, nascimento2024data, zalandmtfgrasp}. Moreover, edge environments are characterized by high volatility, where device availability, network conditions, and resource capacities fluctuate dynamically due to factors such as mobility, energy limitations, and varying workloads \cite{ilager2025decentralized, awaysheh2023big}. Thus, in real-world FL, a subset of edge devices participates in each training round, usually through a \textit{client-selection} algorithm.

A critical aspect that has received limited attention in volatile edge environments is the fairness of client selection algorithms in FL. Most existing FL algorithms prioritize model accuracy and convergence speed, often leading to the repeated selection of high-performing clients while marginalizing devices with limited resources \cite{cho2020client}. This bias not only undermines the democratic principles of federated learning but also creates sustainability issues, as consistently selected clients may experience faster battery depletion and increased wear, potentially leading to their premature exit from the federation \cite{mohri2019agnostic}. On the other hand, recent \textit{fair client selection} algorithms, which consider fairness as the primary objective for client selection in each round, often overlook final model performance and training time. 

The trade-off between accuracy, training speed, and fairness in federated learning becomes particularly pronounced in volatile edge environments. Traditional greedy selection strategies that prioritize clients with superior computational or communication capabilities may achieve faster convergence and higher accuracy in the short term. However, such approaches can lead to significant fairness violations, where a small subset of well-resourced clients dominates the training process while the majority of edge devices remain underutilized \cite{huang2020loadaboost}. Additionally, existing approaches typically consider a fixed resource budget (static resources), whereas in volatile edge environments, device resources can change dramatically between rounds. This raises a fundamental question that motivates our research: \textbf{Is fairness worth the accuracy trade-off in highly volatile edge environments?}

To address the fundamental question, this work presents an extensive empirical evaluation of traditional, greedy, and fair client selection approaches on three benchmarking datasets (CIFAR10, FashionMNIST, and EMNIST), examining the trade-offs between accuracy and fairness, as well as speed and fairness. The experiments consider both static and volatile resource scenarios, and more importantly, consider both independent and identically distributed (IID) and non-IID scenarios \cite{haller2023handling}.

\vspace{.1cm}

The extensive empirical evaluation demonstrates that achieving a trade-off between performance, fairness, and speed, although challenging, can be managed by prioritizing certain desired features in edge environments. While greedy algorithms can achieve up to 25\% improvement in training time, they can suffer from lower fairness and performance, while random strategies can boost higher fairness and competitive performance, but suffer from higher time budgets. While fair client selection strategies try to navigate all three at the same time, they sometimes fail to achieve one of the three. 

\section{Related Work}

\subsection{Federated Learning in Edge Computing}

The convergence of federated learning and edge computing has attracted significant research attention due to the complementary nature of these paradigms. Recent work by Ilager et al. \cite{ilager2025decentralized} addresses the challenges of monitoring volatile edge environments, highlighting the fundamental challenges of managing resources in failure-prone, unsophisticated computing environments \cite{caderno2025bigopera}.

Cross-device federated learning specifically addresses scenarios with volatile connectivity and limited computing resources, such as mobile phones and IoT devices \cite{bonawitz2019towards}. This type of federated learning is particularly relevant to our work, as it explicitly acknowledges the challenges posed by device heterogeneity and intermittent availability that characterize edge environments and in the cloud-to-edge continuum \cite{awaysheh2022cloud}.

\subsection{Fairness in Federated Learning}

The intersection of fairness and federated learning has emerged as a critical area of research\cite{salazar2024survey}. Recent work on quantifying fairness in federated learning has proposed Federated Fairness Analytics, a methodology for measuring fairness that comprises four distinct notions with corresponding metrics \cite{arxiv2024quantifying}. The Self-aware Fairness Federated Learning (SFFL) framework \cite{zhang2025sffl} represents a significant advancement in jointly improving fairness and performance under heterogeneous data distributions. Among others, selection fairness is essential in fair federated learning\cite{shi2023towards}.

\subsection{Fair Client Selection Strategies}

FL, at its core, is built to select a subset of devices in each communication round. Most of the works focus on unbiased with replacement sampling of clients in each round, which helps FL algorithms to theoretically converge similarly to vanilla SGD\cite{cho2022towards}. Since then, however, several biased client selection strategies have been proposed considering the non-IID nature of data across clients in FL. AUCTION\cite{deng2021auction} embeds client selection into an attention-based encoder–decoder network trained via reinforcement learning to automatically learn quality-aware selection policies that balance data size, data quality, and budget constraints for efficient federated learning. Tan et al. \cite{tan2022reputation} represents a first-of-its-kind reputation-aware approach that optimally selects and compensates clients with different reputation profiles. FedCo\cite{tang2022fedcor} embeds a Gaussian Process model of inter-client loss correlations into federated learning, deriving a correlation-driven client selection policy. 

Recent works have also incorporated fairness into the client selection process to make the selection process more equitable. FairFedCS\cite{shi2023fairness}, a Lyapunov optimization–driven client selection framework, dynamically adjusts each client’s selection probability by jointly accounting for reputation, past participation, and performance contribution. RBCS-F\cite{huang2020efficiency} estimates per-client exchange rates and maintains long-term fairness via virtual queues to prioritize clients with lesser involvement over time. 

\section{Preliminaries}

\subsection{Federated Learning Framework}

We consider a federated learning system comprising $N$ edge clients, $\mathcal{C} = \{1, 2, \ldots, N\}$, and a central server. Each client $i \in \mathcal{C}$ possesses a local dataset $D_i$ containing $|D_i|$ samples. The global objective is to train a machine learning model that minimizes: collaboratively

\begin{equation}
F(w) = \sum_{i=1}^N \frac{|D_i|}{|D|} F_i(w)
\end{equation}

where $F_i(w) = \frac{1}{|D_i|} \sum_{(x,y) \in D_i} \ell(w; x, y)$ represents the local loss function for client $i$.

\subsection{Volatile Edge Environment Model}

Edge environments are characterized by high volatility in terms of resource availability. We model this volatility through dynamic resource capabilities:

\begin{align}
c_i^{comp}(t) &\sim \mathcal{U}(c_{min}^{comp}, c_{max}^{comp}) \\
c_i^{comm}(t) &\sim \mathcal{U}(c_{min}^{comm}, c_{max}^{comm})
\end{align}

where $\mathcal{U}(a,b)$ denotes the uniform distribution over interval $[a,b]$.

\subsection{Fairness Formulation}

We adopt Jain's Fairness Index (JFI) as our primary fairness metric:

\begin{equation}
\text{JFI}(s) = \frac{(\sum_{i=1}^N s_i)^2}{N \cdot \sum_{i=1}^N s_i^2}
\end{equation}

where $s_i$ represents the number of times client $i$ has been selected for participation.

\subsection{Reputation-Based Fair Federated Learning (RBFF)\cite{song2021reputation}}

RBFF incorporates both client capabilities and fairness considerations through a reputation-based scoring mechanism.

\textbf{Reputation Computation:} At each round $t$, we compute:
\begin{equation}
r_i^t = c_i^{comp}(t) + c_i^{comm}(t)
\end{equation}

\textbf{Fairness-Aware Selection Score:}
\begin{equation}
\text{s}_i^t = \frac{r_i^t}{1 + s_i^{t-1}}
\end{equation}

\subsection{Reputation-Based Client Selection with Fairness (RBCSF)}

RBCSF provides stronger fairness guarantees through direct penalization:

\begin{equation}
\text{s}_i^{t,RBCSF} = r_i^t - \alpha \cdot s_i^{t-1}
\end{equation}

where $\alpha > 0$ controls the strength of the fairness penalty.

\section{Experimental Framework}

\subsection{Experimental Setup}

We conduct comprehensive experiments using three widely-used benchmark datasets that represent different complexity levels and application domains. Our experimental framework is designed to evaluate the performance of traditional client selection strategies against state-of-the-art fair client selection strategies.

\textbf{Datasets and Models:} Our evaluation encompasses three datasets:
\begin{itemize}
\item \textit{EMNIST (Extended MNIST):} A character recognition dataset containing 814,255 samples across 62 classes
\item \textit{FashionMNIST:} A fashion product classification dataset with 70,000 grayscale images across 10 clothing categories.
\item \textit{CIFAR-10:} A natural image classification dataset containing 60,000 color images across 10 object classes. 
\end{itemize}
We employ a CNN architecture with two convolutional layers (32 and 64 filters), followed by max pooling and two fully connected layers (128 and output neurons).

\textbf{Federation Configurations:} We evaluate scalability across different federation sizes, 50 clients if not mentioned otherwise. The configuration represents a scale of edge deployments, from small to medium-sized clusters.

\textbf{Data Distribution Scenarios:} To capture the heterogeneity challenges in real federated learning deployments, we implement three data distribution models:
\begin{itemize}
\item \textit{IID (Independent and Identically Distributed):} Data is uniformly distributed across clients, representing the ideal scenario.
\item \textit{Class Non-IID:} Each client has data from only 2 classes out of the total available classes, creating significant statistical heterogeneity.
\item \textit{Quantity Skew:} Each client receives a specific portion of the data calculated from a Dirichlet distribution. 
\end{itemize}

\textbf{Volatility Modeling:} We implement two environmental scenarios:
\begin{itemize}
\item \textit{Static Environment:} Client capabilities remain constant throughout training.
\item \textit{Dynamic Environment:} Client computational and communication capabilities change randomly each round according to: $c_i^{comp}(t) \sim \mathcal{U}(50, 200)$ samples/second and $c_i^{comm}(t) \sim \mathcal{U}(10^5, 5 \times 10^5)$ bits/second.
\end{itemize}

\textbf{Training Configuration:} All experiments use consistent parameters: local epochs $E = 1$, learning rate $\eta = 0.01$, batch size 32, total rounds $T = 50$, and client selection ratio 40\%, unless stated otherwise.

\subsection{Baseline Algorithms}

The following baseline client selection algorithms have been considered in this work:
\begin{itemize}
\item \textit{Random Selection:} Uniform random client selection providing a fairness baseline.
\item \textit{Computational Greedy (Comp\_Greedy):} Prioritizes clients with highest computational speeds.
\item \textit{Communication Greedy (Comm\_Greedy):} Selects clients with highest communication speeds.
\item \textit{RBFF:} Incorporates both client capabilities and fairness considerations in client selection.
\item \textit{RBCSF:} Adds to RBFF with penalization.
\end{itemize}

\subsection{Metrics}
In this work, we report the accuracy of the global model on the test set, alongside JFI, a fairness metric. We also report the area under the curve (AUC) and the receiver operating characteristic (ROC) curve, along with the wall time. Please note that the wall time is reported in thousand seconds.

\section{Comparative Analysis Results}

\begin{table*}[htbp]
  \centering
  \caption{Performance of different methods under various data distributions on the EMNIST dataset with 50 clients.}
  \label{tab:emnist}
  \begin{tabularx}{\textwidth}{m{1.2cm}m{1.8cm}*{15}{X}}
    \toprule
    Resource & Method & \multicolumn{5}{c}{IID} & \multicolumn{5}{c}{Class non-IID} & \multicolumn{5}{c}{Quantity Skew} \\
    \cmidrule(lr){3-7} \cmidrule(lr){8-12} \cmidrule(lr){13-17}
      &  & Acc $\uparrow$ & Time $\downarrow$ & JFI $\uparrow$ & AUC $\uparrow$ & ROC $\uparrow$ &   Acc $\uparrow$ & Time $\downarrow$ & JFI $\uparrow$ & AUC $\uparrow$ & ROC $\uparrow$   & Acc $\uparrow$ & Time $\downarrow$ & JFI $\uparrow$ & AUC $\uparrow$ & ROC $\uparrow$  \\  
    \midrule
    \multirow{5}{*}{Static}
      & Random       & 75.76 & 28.79 & \textbf{0.980} & 0.990 & 0.991 & 31.61 & 25.49 & \textbf{0.973} & 0.894 & 0.863 & 76.25 & 25.91 & \textbf{0.973} & 0.991 & 0.992 \\
      & RBFF         & \textbf{78.03} & 26.06 & 0.885 & 0.992 & 0.993 & \textbf{35.03} & 24.02 & 0.873 & 0.890 & 0.855 & 76.59 & 28.92 & 0.883 & 0.991 & 0.992 \\
      & RBCSF        & 76.75 & 24.24 & 0.776 & 0.991 & 0.992 & 29.85 & 26.56 & 0.726 & 0.893 & 0.860 & \textbf{77.00} & 23.74 & 0.742 & 0.991 & 0.992 \\
      & Comm-greedy  & 76.23 & 29.26 & 0.400 & 0.990 & 0.991 & 28.61 & 20.98 & 0.400 & 0.770 & 0.688 & 75.33 & 17.71 & 0.400 & 0.989 & 0.990 \\
      & Comp-greedy  & 77.72 & \textbf{19.35} & 0.400 & 0.991 & 0.992 & 29.70 & \textbf{17.53} & 0.400 & 0.754 & 0.679 & 73.50 & \textbf{16.67} & 0.400 & 0.988 & 0.989 \\
    \midrule
    \multirow{5}{*}{Volatile}
      & Random       & 77.80 & 27.80 & 0.969 & 0.991 & 0.992 & 38.19 & 26.26 & 0.969 & 0.912 & 0.866 & 76.28 & 25.86 & 0.971 & 0.990 & 0.991 \\
      & RBFF         & 77.16 & 24.65 & \textbf{0.993} & 0.991 & 0.992 & \textbf{38.57} & 22.92 & \textbf{0.994} & \textbf{0.915} & \textbf{0.875} & 77.08 & 24.61 & \textbf{0.996} & 0.991 & 0.992 \\
      & RBCSF        & 76.35 & 25.17 & 0.986 & 0.990 & 0.991 & 37.86 & 22.33 & 0.989 & 0.904 & 0.870 & 76.99 & 24.48 & 0.990 & 0.991 & 0.992 \\
      & Comm-greedy  & \textbf{78.07} & 25.19 & 0.964 & \textbf{0.992} & \textbf{0.993} & 36.87 & 22.98 & 0.964 & 0.904 & 0.870 & 77.37 & 24.23& 0.956 & \textbf{0.992} & 0.992 \\
      & Comp-greedy  & 77.22 & \textbf{20.41} & 0.976 & 0.991 & 0.992 & 36.26 & \textbf{19.18} & 0.973 & 0.911 & 0.871 & \textbf{77.89} & \textbf{20.12} & 0.968 & 0.991 &\textbf{0.993} \\
    \bottomrule
  \end{tabularx}
\end{table*}

\begin{table*}[htbp]
  \centering
  \caption{Performance of different methods under various data distributions on the FashionMNIST dataset with 50 clients.}
  \label{tab:femnist}
  \begin{tabularx}{\textwidth}{m{1.2cm} m{1.8cm} *{15}{X}}
    \toprule
    Resource & Method & \multicolumn{5}{c}{IID} & \multicolumn{5}{c}{Class non-IID} & \multicolumn{5}{c}{Quantity Skew} \\
    \cmidrule(lr){3-7} \cmidrule(lr){8-12} \cmidrule(lr){13-17}
      & & Acc $\uparrow$ & Time $\downarrow$ & JFI $\uparrow$ & AUC $\uparrow$ & ROC $\uparrow$  &  Acc $\uparrow$ & Time $\downarrow$ & JFI $\uparrow$ & AUC $\uparrow$ & ROC $\uparrow$  &  Acc $\uparrow$ & Time $\downarrow$ & JFI $\uparrow$ & AUC $\uparrow$ & ROC $\uparrow$  \\  
    \midrule
    \multirow{5}{*}{Static}
      & Random       & 80.05 & 18.98 & \textbf{0.963} & 0.976 & 0.983 & 68.10 & 18.53 & \textbf{0.970} & 0.957 & 0.959 & \textbf{80.57} & 21.64 & \textbf{0.976} & \textbf{0.977} & \textbf{0.984} \\
      & RBFF         & \textbf{80.54} & 17.57 & 0.848 & \textbf{0.977} & \textbf{0.984} & \textbf{68.88} & 17.06 & 0.875 & 0.959 & \textbf{0.964} & 80.40 & 17.57 & 0.824 & 0.976 & \textbf{0.984} \\
      & RBCSF        & 80.31 & 17.18 & 0.716 & 0.976 & 0.983 & 61.68 & 17.10 & 0.775 & 0.954 & 0.952 & 80.01 & 18.80 & 0.748 & \textbf{0.977} & \textbf{0.984} \\
      & Comm-greedy  & 80.33 & 15.14 & 0.400 & \textbf{0.977} & \textbf{0.984} & 63.27 & 16.24 & 0.400 & 0.950 & 0.935 & 80.30 & 17.46& 0.400 & 0.976 & \textbf{0.984} \\
      & Comp-greedy  & 80.51 & \textbf{13.24} & 0.400 & 0.976 & \textbf{0.984} & 61.62 & \textbf{13.38} & 0.400 & \textbf{0.960} & 0.954 & 78.47 & \textbf{10.84} & 0.400 & 0.973 & 0.981 \\
    \midrule
    \multirow{5}{*}{Volatile}
      & Random       & 79.84 & 17.92 & 0.979 & 0.976 & 0.983 & 63.36 & 17.64 & 0.976 & 0.960 & 0.953 & 79.88 & 17.63 & 0.978 & 0.976 & 0.983 \\
      & RBFF         & 80.19 & 15.20 & \textbf{0.994} & 0.976 & 0.983 & 62.30 & 15.01 & \textbf{0.992} & 0.960 & 0.951 & 79.93 & 15.55 & \textbf{0.994} & 0.976 & 0.983 \\
      & RBCSF        & 80.10 & 15.08 & 0.991 & 0.977 & 0.984 & \textbf{66.92} & 15.50 & 0.991 & \textbf{0.961} & \textbf{0.960} & 80.04 & 15.44 & 0.989 & 0.976 & 0.983 \\
      & Comm-greedy  & \textbf{81.11} & 14.49 & 0.970 & 0.977 & 0.985 & 60.85 & 15.23 & 0.973 & 0.956 & 0.954 & \textbf{80.57} & 15.20 & 0.974 & 0.977 & \textbf{0.984} \\
      & Comp-greedy  & 80.54 & \textbf{13.96} & 0.970 & 0.977 & 0.984 & 63.51 & \textbf{13.99} & 0.960 & 0.957 & 0.953 & 79.93 & \textbf{14.32} & 0.967 & 0.976 & 0.983 \\
    \bottomrule
  \end{tabularx}
\end{table*}

\begin{table*}[!h]
  \centering
  \caption{Performance of different methods under various data distributions on the CIFAR10 dataset with 50 clients.}
  \label{tab:cifar10}
  \begin{tabularx}{\textwidth}{m{1.2cm} m{1.8cm} *{15}{X}}
    \toprule
    Resource & Method & \multicolumn{5}{c}{IID} & \multicolumn{5}{c}{Class non-IID} & \multicolumn{5}{c}{Quantity Skew} \\
    \cmidrule(lr){3-7} \cmidrule(lr){8-12} \cmidrule(lr){13-17}
      &  & Acc $\uparrow$ & Time $\downarrow$ & JFI $\uparrow$ & AUC $\uparrow$ & ROC $\uparrow$  &  Acc $\uparrow$ & Time $\downarrow$ & JFI $\uparrow$ & AUC $\uparrow$ & ROC $\uparrow$  &  Acc $\uparrow$ & Time $\downarrow$ & JFI $\uparrow$ & AUC $\uparrow$ & ROC $\uparrow$  \\  
    \midrule
    \multirow{5}{*}{Static}
      & Random       & 35.69 & 18.05 & \textbf{0.964} & 0.797 & 0.799 & 18.86 & 18.60 & \textbf{0.968} & \textbf{0.743} & \textbf{0.687} & 36.27 & 19.87 & \textbf{0.967} & 0.800 & 0.804 \\
      & RBFF         & 36.40 & 15.97 & 0.901 & 0.801 & 0.804 & 20.74 & 15.77 & 0.878 & 0.741 & 0.675 & 35.53 & 17.87 & 0.829 & 0.793 & 0.792 \\
      & RBCSF        & 35.97 & 16.16 & 0.734 & 0.794 & 0.798 & 24.38 & \textbf{15.20} & 0.764 & 0.734 & 0.670 & 37.35 & 17.18 & 0.731 & 0.808 & 0.810 \\
      & Comm-greedy  & \textbf{37.44} & \textbf{14.22} & 0.400 & \textbf{0.804} & \textbf{0.808} & \textbf{24.42} & 16.22 & 0.400 & 0.736 & 0.681 & 34.03 & \textbf{16.72} & 0.400 & 0.782 & 0.785 \\
      & Comp-greedy  & 36.34 & 14.95 & 0.400 & 0.798 & 0.800 & 16.19 & 15.35 & 0.400 & 0.723 & 0.637 & \textbf{38.35} & 17.63 & 0.400 & \textbf{0.814} & \textbf{0.817} \\
    \midrule
    \multirow{5}{*}{Volatile}
      & Random       & 35.33 & 17.88 & 0.971 & 0.788 & 0.791 & 23.65 & 18.07 & 0.967 & 0.749 & 0.717 & 35.19 & 18.14 & 0.972 & 0.797 & 0.798 \\
      & RBFF         & 35.22 & 14.72 & \textbf{0.993} & 0.788 & 0.792 & 23.61 & 14.67 & \textbf{0.994} & 0.747 & \textbf{0.726} & 33.74 & 14.53& \textbf{0.994} & 0.789 & 0.792 \\
      & RBCSF        & 35.66 & 14.67 & 0.989 & \textbf{0.793} & \textbf{0.797} & 23.28 & 14.52 & 0.991 & \textbf{0.753} & 0.710 & \textbf{35.85} & \textbf{14.13} & 0.991 & 0.795 & 0.800 \\
      & Comm-greedy  & \textbf{35.96} & \textbf{14.44} & 0.974 & 0.792 & 0.796 & \textbf{24.20} & \textbf{14.41} & 0.968 & 0.752 & 0.690 & 35.84 & 14.43 & 0.977 & \textbf{0.801} & \textbf{0.802} \\
      & Comp-greedy  & 35.29 & 14.46 & 0.977 & 0.786 & 0.791 & 21.85 & 14.81 & 0.969 & 0.743 & 0.708 & 35.25 & 14.44 & 0.979 & 0.791 & 0.793 \\
    \bottomrule
  \end{tabularx}
\end{table*}

\subsection{Main Results}
The main results of the comparative analysis are arranged in individual tables for each dataset. Table \ref{tab:emnist} for EMNIST, Table \ref{tab:femnist} for FashionMNIST, and Table \ref{tab:cifar10} for the CIFAR10 dataset. Our comprehensive experimental evaluation addresses the central research question: \textbf{Is fairness worth the accuracy trade-off in highly volatile edge environments?}

In experiments where clients own \textbf{static resources}, it is noticeable that random client selection has superior fairness overall in all three datasets. Random client selection overperforms both greedy and fairness-based client selection algorithms, with around \textbf{10\%} overall boost in the JFI score. It also achieves within \textbf{1-2\%} in absolute terms, compared to the best-performing client selection algorithm in terms of performance, as reported by accuracy. The fairness-based client selection algorithms achieve closer performance in terms of fairness to random selection in static resource environments, but yield higher performance and lower wall time in most experiments. The greedy client selection algorithms, specifically computation-based greedy client selection, achieve the best overall wall-clock time for the experiments, as the best-performing clients are selected repeatedly in each round of the experiments. This, however, results in a JFI score proportional to the portion of clients chosen in each round, which is set to 0.4, representing the lowest JFI. The time improvement in greedy algorithms in the static resource experiments, however, is enormous, and can amount to \textbf{25\%} higher than other client selection strategies. However, greedy client selection suffers from lower performance, which in some cases can be as high as \textbf{2-3\%} in absolute terms. The outlier, however, is the experiments on the \textbf{CIFAR-10} dataset, where greedy algorithms boost lower clock time and higher accuracy overall.

\begin{figure*}
    \centering
    \includegraphics[width=0.95\linewidth]{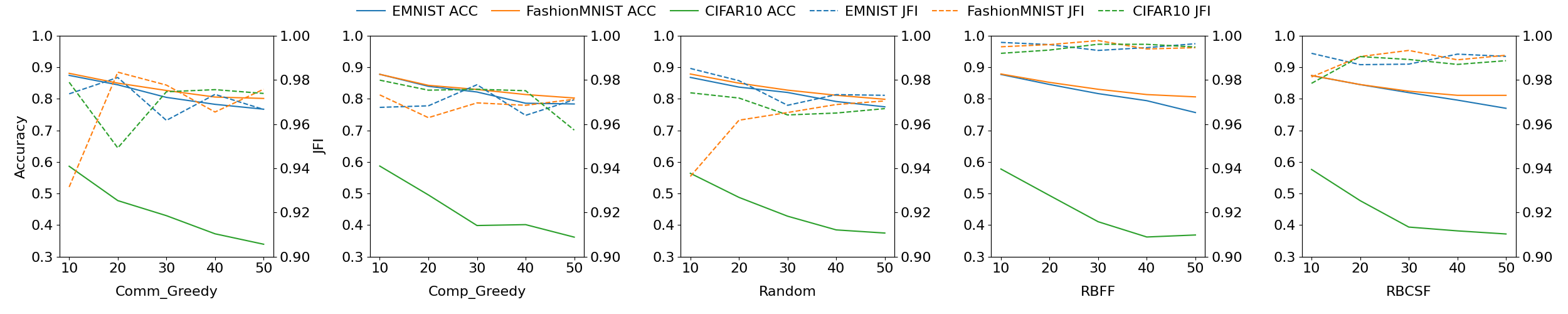}
    \caption{Impact of number of clients on accuracy JFI tradeoff under volatile resources}
    \label{fig:dynamic_jfi_acc_clients}
\end{figure*}

\begin{figure*}
    \centering
    \includegraphics[width=\linewidth]{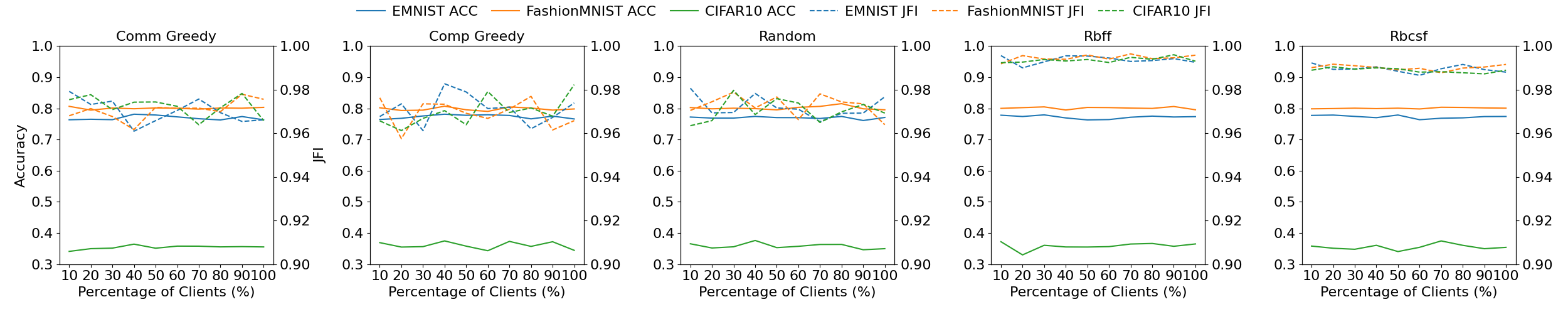}
    \caption{Impact of percentage of selected clients on accuracy JFI tradeoff under volatile resources}
    \label{fig:dynamic_jfi_acc_percentage}
\end{figure*}

\subsection{Volatility Impact Analysis}

In environments where client resources are volatile, the equations change from those in static environments. The random client selection algorithm is not the \textit{fairest} algorithm anymore. The RBFF client selection algorithms outperform all other approaches in all experiments, consistently achieving higher JFI values in each experiment. The equation for greedy client selection algorithms remains similar, mainly in static environments. Greedy client selection algorithms yield the lowest overall clock time (communication greedy client selection for CIFAR10 and computation greedy client selection for the other two datasets). The wall clock time gains remain in the same region, \textbf{20-25\%} in percentage terms, except for the CIFAR-10 dataset experiments, where the gains are minimal.

\subsection{Impact of Scale and Participation}

In this section, we discuss the impact of scale and participation rate on the accuracy and fairness trade-off that arises from increasing the number of participating edge devices. Figure \ref{fig:dynamic_jfi_acc_clients} illustrates the accuracy and JFI score for varying numbers of clients, ranging from 10 to 50. In most cases, while the accuracy drops as a result of lower samples at each client, the JFI scores remain mostly intact overall. Furthermore, Figure \ref{fig:dynamic_jfi_acc_percentage} shows the impact of increasing participation rate in each round on the performance and JFI score. While the JFI fluctuates for the greedy algorithms, it remains at high levels for the rest of the algorithms across all participation rates. The accuracy, although fluctuating for all algorithms, remains relatively consistent overall. 

\subsection{Impact of Data Heterogeneity}

\begin{figure*}[htbp]
\centering
\includegraphics[width=\textwidth]{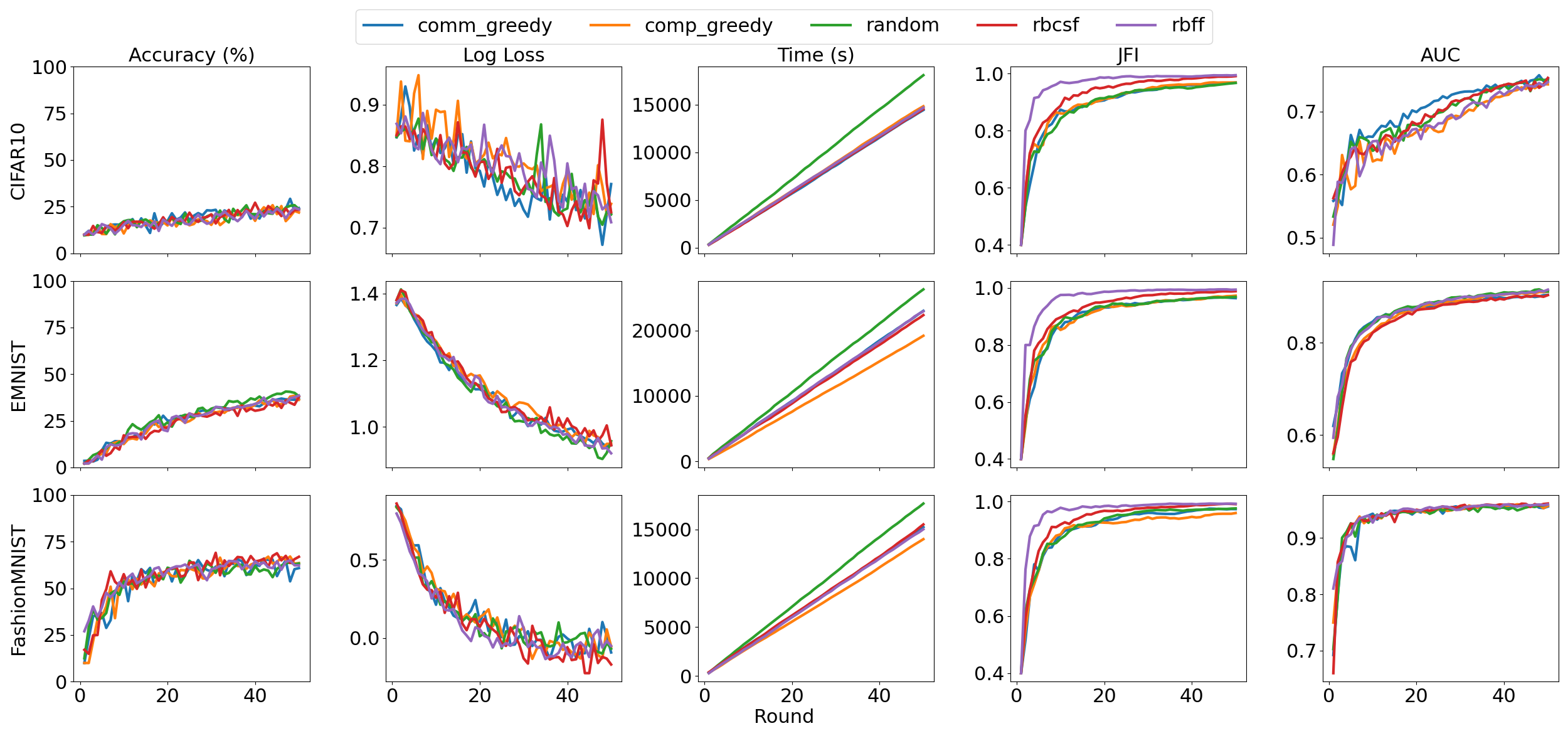}
\caption{Performance comparison in highly volatile edge environments with statistical heterogeneity (Class Non-IID, 50 clients, dynamic). This scenario represents the intersection of edge intelligence constraints, high volatility, and maximum fairness challenges. RBFF demonstrates superior fairness (JFI = 0.99X) while maintaining competitive accuracy in the most challenging experimental condition.}
\label{fig:class_noniid_50_dynamic}
\end{figure*}

The results in Figure~\ref{fig:class_noniid_50_dynamic} reveal several critical insights. While the log loss fluctuation is normal as a portion of \textit{different} clients are selected in each training round, and the time linearly improves for all the experiments, the JFI progression is different for each algorithm. RBFF exceptionally achieves and maintains the highest JFI score, boosting fairness from the initial few rounds, while other algorithms, such as RBCSF and random client selection algorithms, catch up to get closer to RBFF; it still maintains the highest fairness. On the accuracy front, there is minimal to no difference between the algorithms across datasets. 

\begin{figure*}[t!]
\centering
\includegraphics[width=\textwidth]{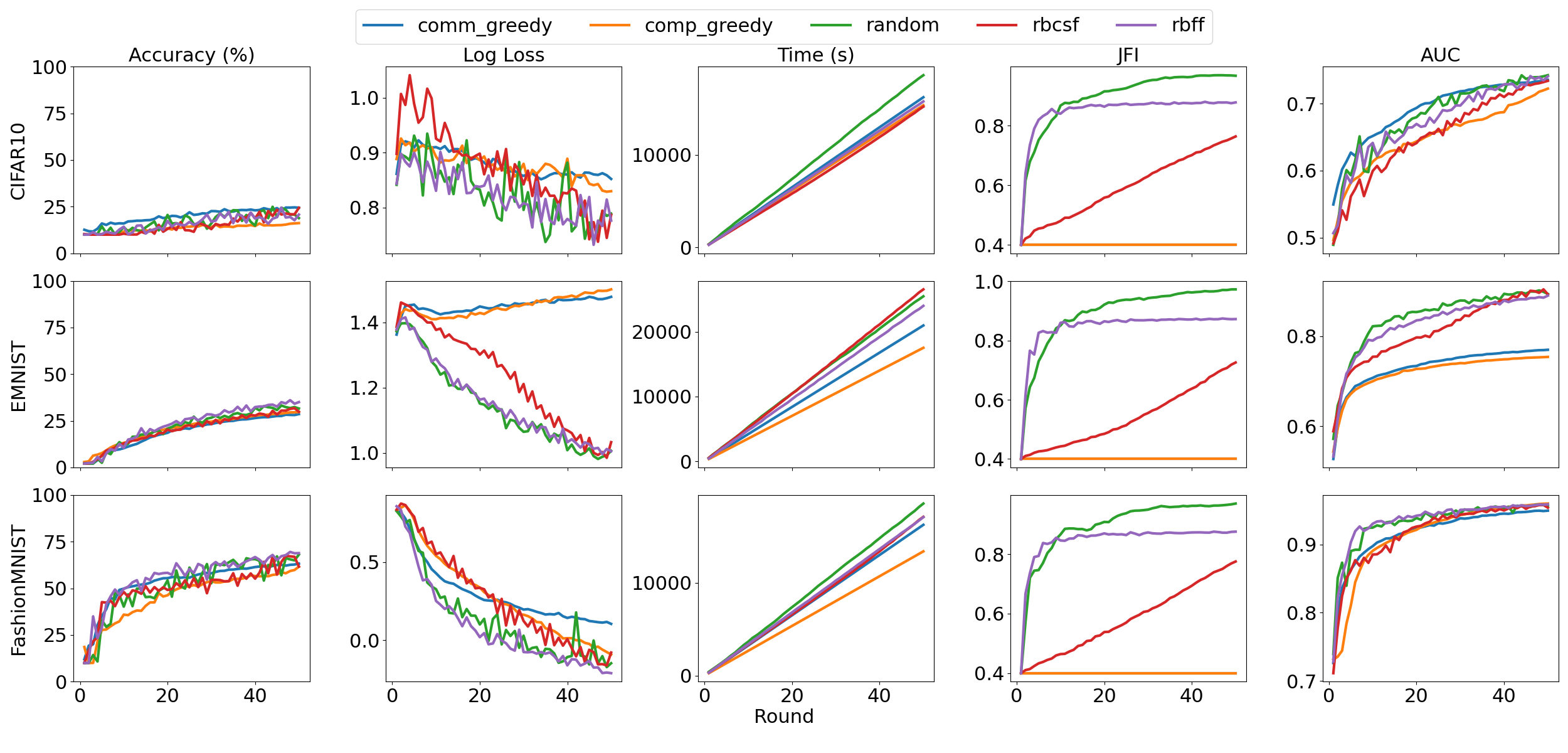}
\caption{Baseline performance in stable edge environments for volatility impact assessment (Class Non-IID, 50 clients, static). Direct comparison with Figure~\ref{fig:class_noniid_50_dynamic} enables precise quantification of volatility effects on fairness-aware client selection algorithms.}
\label{fig:class_noniid_50_static}
\end{figure*}

To quantify the specific impact of volatility, Figure~\ref{fig:class_noniid_50_static} provides the essential baseline comparison under identical conditions but in a stable environment.

The comparison between Figures~\ref{fig:class_noniid_50_dynamic} and~\ref{fig:class_noniid_50_static} provides unprecedented insight into volatility effects. Remarkably, the dynamic environment actually enhances the relative benefits of fairness-aware selection. This counterintuitive finding aligns with recent work on adaptive federated learning and suggests that volatility, rather than being purely detrimental, creates conditions where fairness mechanisms provide additional robustness benefits. The stable JFI maintenance in dynamic conditions demonstrates the inherent adaptability of reputation-based client selection.

\subsection{Data Quantity Heterogeneity Impact Assessment}

\begin{figure*}[t!]
\centering
\includegraphics[width=\textwidth]{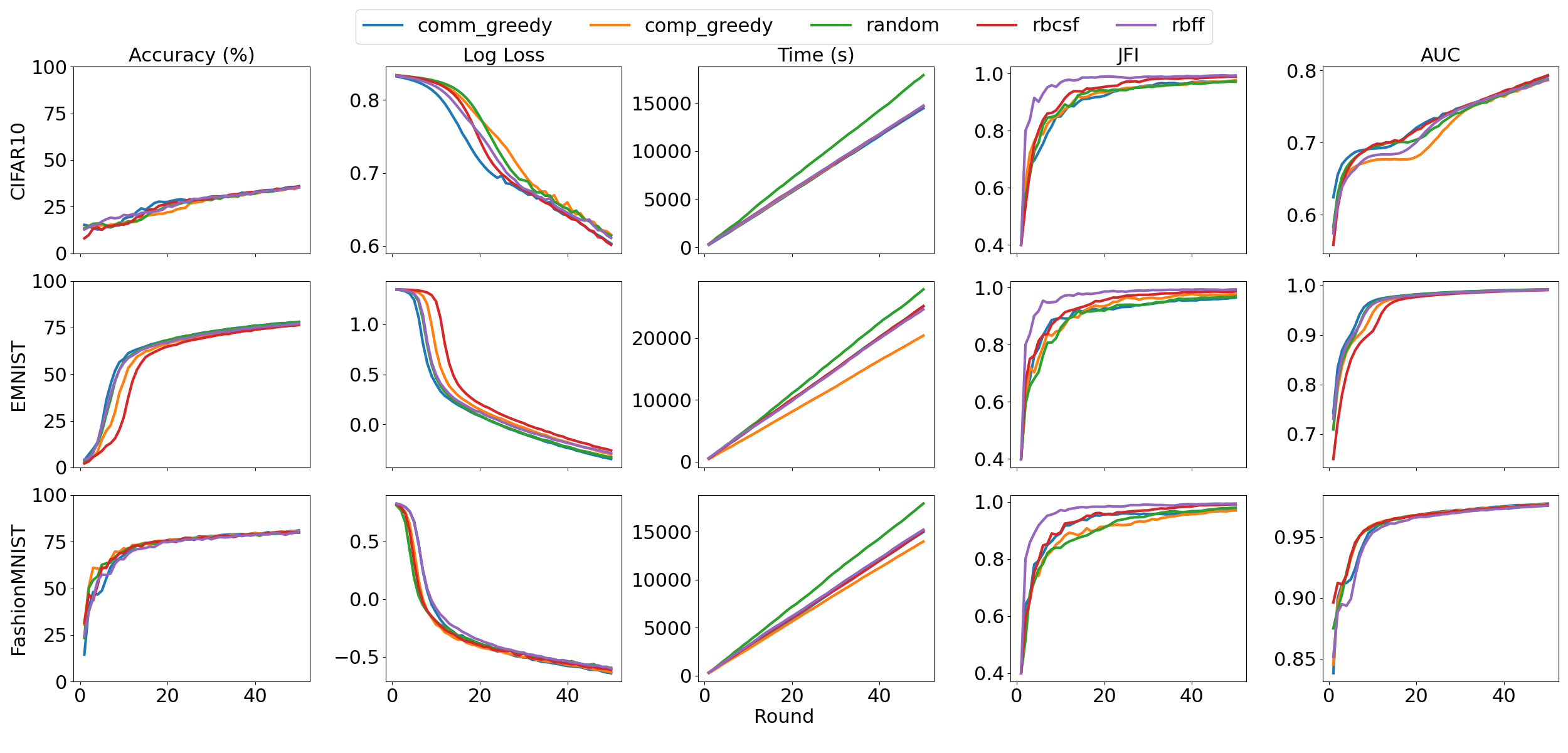}
\caption{Fairness algorithm performance under uniform data distribution in volatile edge environments (IID, 50 clients, dynamic). This scenario illustrates optimal conditions for fairness mechanisms, demonstrating how algorithms adapt when statistical heterogeneity is minimized while environmental volatility persists.}
\label{fig:iid_50_dynamic}
\end{figure*}

Understanding how fairness algorithms perform across different types of data heterogeneity is crucial for practical deployment. Figure~\ref{fig:iid_50_dynamic} examines algorithm behavior under uniform data distribution in volatile environments.

Figure~\ref{fig:iid_50_dynamic} reveals the upper bound of algorithm performance when data distribution challenges are minimized. RBFF achieves exceptional results with JFI = 0.993 and accuracy = (36, 81, and 78)\%, demonstrating that fairness mechanisms excel when statistical heterogeneity is reduced. 

The contrast with Figure~\ref{fig:class_noniid_50_dynamic} illustrates the algorithm's adaptability across the heterogeneity spectrum. The 13\% accuracy improvement (from 24\% to 37\%) when moving from class non-IID to IID conditions in the CIFAR10 dataset demonstrates the expected impact of data distribution. At the same time, the consistently high fairness (JFI difference of only 0.04) proves the algorithm's robustness across scenarios.

\vspace{.4cm}

\begin{figure*}[t!]
\centering
\includegraphics[width=\textwidth]{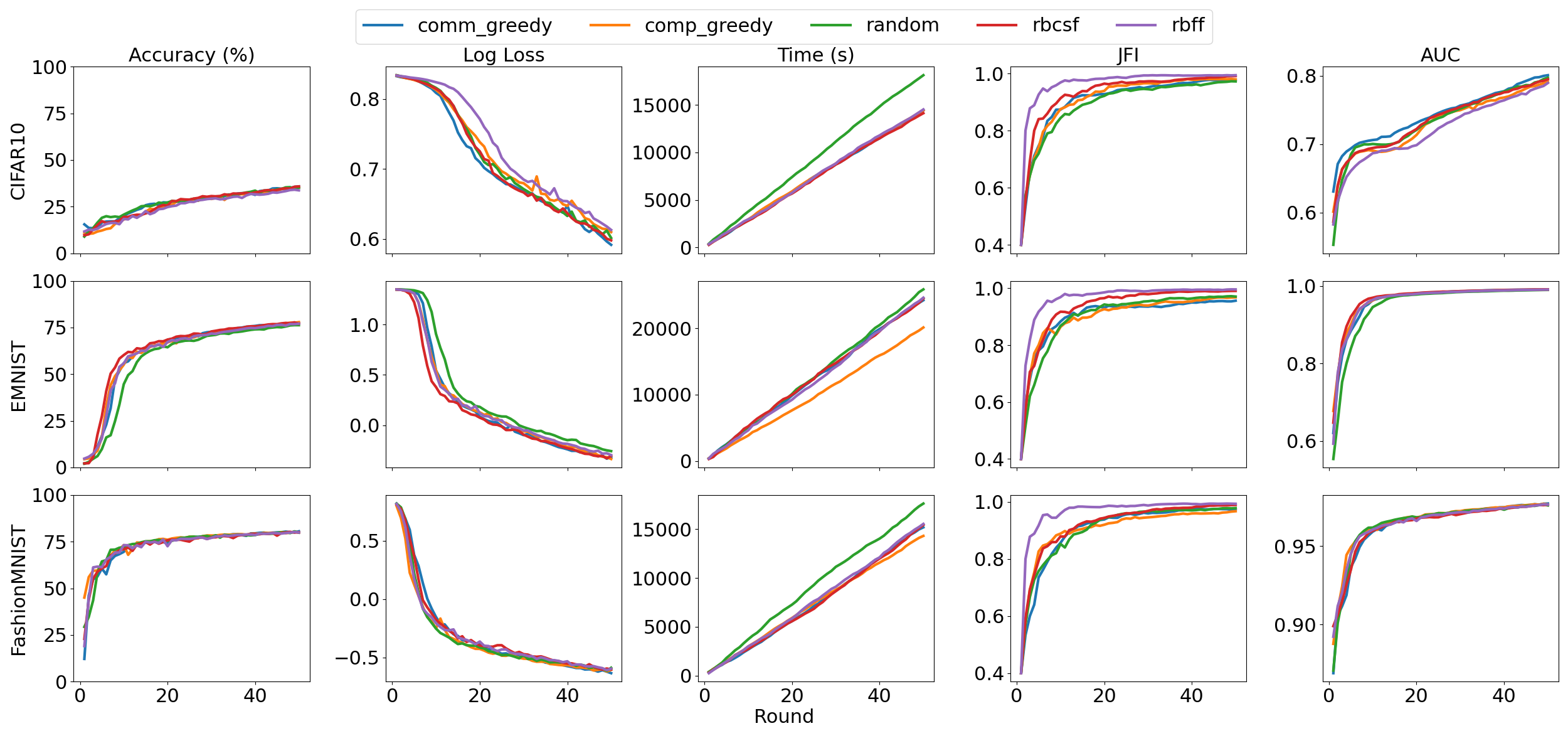}
\caption{Impact of system heterogeneity on fairness-aware client selection in volatile environments (Quantity Skew, 50 clients, dynamic). This scenario illustrates algorithm performance under varying data volumes, representing common edge deployment conditions where device capabilities and data collection rates differ significantly.}
\label{fig:quantity_skew_50_dynamic}
\end{figure*}

To complete the heterogeneity analysis, Figure~\ref{fig:quantity_skew_50_dynamic} examines system heterogeneity effects through quantity skew scenarios.

Figure~\ref{fig:quantity_skew_50_dynamic} provides crucial insights into system heterogeneity effects, where clients possess different data volumes rather than different data distributions. The results demonstrate intermediate performance characteristics between IID and class non-IID scenarios, with accuracy and fairness metrics that reflect the moderate challenge level posed by quantity skew. 

The quantity skew results demonstrate that RBFF maintains robust fairness performance (JFI $>$ 0.97) while achieving competitive accuracy. This consistency across heterogeneity types validates the algorithm's practical applicability in diverse real-world scenarios where both statistical and system heterogeneity may coexist.

This scenario examines how system heterogeneity—particularly quantity skew across 50 dynamic clients—affects fairness-aware client selection, reflecting real-world edge environments with diverse device capabilities and data generation rates.

\section{Conclusion}

In this work, we presented a comprehensive empirical investigation of client selection strategies in federated learning over volatile edge environments, evaluating traditional, greedy, random, and fairness-aware approaches on CIFAR-10, FashionMNIST, and EMNIST under both IID and non-IID data distributions. Our results reveal that, contrary to common belief, fairness and high model performance need not be mutually exclusive. By integrating measured fairness constraints into the selection process, we can achieve substantial gains in client participation equity without incurring prohibitive losses in accuracy or convergence speed. In highly dynamic settings—where device availability, energy levels, and network conditions fluctuate—naïve greedy strategies tend to exacerbate participation imbalances over time. In contrast, fairness-aware methods not only stabilize client inclusion but also prolong the effective lifespan of resource-constrained devices.

Beyond these findings, our study underscores the importance of configurability in practical deployments: by tuning the relative emphasis on accuracy, speed, and fairness, practitioners can tailor the client selection mechanism to their application’s specific requirements. Looking ahead, integrating formal privacy guarantees, such as differential privacy, directly into the fairness-aware selection pipeline would ensure that equity enhancements do not compromise user confidentiality. Furthermore, extending fairness notions beyond per-round participation to cross-round or cumulative metrics would provide stronger, long-term assurances of equitable involvement.

\balance

\bibliography{main}

\end{document}